\documentclass[sigconf]{acmart}
\usepackage{subcaption} 

\usepackage{multirow}
\usepackage{enumitem}
\usepackage{pifont}
\usepackage{makecell}
\usepackage[table]{xcolor}

\definecolor{lightcyan}{rgb}{0.88, 1.0, 1.0}
\definecolor{aliceblue}{rgb}{0.94, 0.97, 1.0}

\usepackage{graphicx}

\AtBeginDocument{%
  }

\author{Fengbin Zhu$^{\ast\spadesuit\dagger}$, Zijing Cai $^{\ast\clubsuit}$, Yuzhe Wang $^{\clubsuit}$, Pengyang Shao$^{\spadesuit}$, \\ 
Wenjie Wang$^{\clubsuit}$, Fuli Feng$^{\clubsuit}$, Richang Hong$^{\heartsuit}$, Tat-Seng Chua$^{\spadesuit}$}
\affiliation{
  \institution{
    $^{\spadesuit}$National University of Singapore, Singapore \\
    $^{\clubsuit}$University of Science and Technology of China, China \\
    $^{\heartsuit}$Hefei University of Technology, China
  }
  \country{} 
}

\thanks{$^{\ast}$\textbf{Equal Contribution}}
\thanks{$^{\dagger}$\textbf{Corresponding Author: Fengbin Zhu, fengbin@nus.edu.sg}}

\renewcommand{\shortauthors}{Zhu, et al.}

\setcopyright{acmlicensed}
\copyrightyear{2018}
\acmYear{2018}
\acmDOI{XXXXXXX.XXXXXXX}
\acmConference[Conference acronym 'XX]{Make sure to enter the correct
  conference title from your rights confirmation email}{June 03--05,
  2018}{Woodstock, NY}
\acmISBN{978-1-4503-XXXX-X/2018/06}

\newcommand{\method}{\textsc{MURE}~}
\newcommand{\paradigm}{\texttt{X-VisEmb}~}

\definecolor{vp}{RGB}{237, 125, 49}
\definecolor{vr}{RGB}{68, 114, 196}

\newcommand{\ie}{\emph{i.e., }}
\newcommand{\eg}{\emph{e.g., }}

\begin{document}

\title{MURE: Hierarchical Multi-Resolution Encoding via Vision-Language Models for Visual Document Retrieval
}


\renewcommand{\shortauthors}{Trovato et al.}

\begin{abstract}
  Visual Document Retrieval (VDR) requires representations that capture both fine-grained visual details and global document structure to ensure retrieval efficacy while maintaining computational efficiency. Existing VDR models struggle to balance effectiveness and efficiency when processing high-resolution documents: they often either lose fine-grained information or generate an excessive number of visual tokens, resulting in significant indexing overhead and high retrieval latency.
  In this work, we rethink the visual encoding mechanism and propose a new \textbf{\paradigm} paradigm that progresses from multi-resolution sampling and encoding, through cross-granularity feature fusion, to adaptive representation distillation. A preliminary study validates its feasibility and effectiveness in capturing complementary visual cues at varying scales.
  Building on the insights, we develop \textbf{\method}, a novel framework that employs VLMs as a hierarchical multi-resolution encoder, integrates resolution-level Matryoshka representation learning (RMRL) for effective feature fusion, and applies a semantic-aware hierarchical clustering mechanism for visual token compression. Experiments on two widely used VDR benchmarks show that our \method framework consistently beats strong baselines. Furthermore, it significantly outperforms ColPali with only 50\% of its visual token budget.
\end{abstract}




\received{20 February 2007}
\received[revised]{12 March 2009}
\received[accepted]{5 June 2009}

\maketitle

\section{Introduction}

Visual Document Retrieval (VDR) aims at searching for the most relevant visually-rich documents from a large-scale repository in response to a textual query~\cite{fayssecolpali, yuvisrag}. 
In contrast to conventional document retrieval relying on lossy Optical Character Recognition (OCR) to transform document pages to linear text~\cite{kim2022ocr, xu2020layoutlm}, VDR treats document pages as holistic visual inputs, wherein the essential layout and visual information is preserved and exploited~\cite{lee2023pix2struct}.
Hence, high-quality visual document representation is very critical for VDR.
Recent research has increasingly used Vision-Language Models (VLMs) as document image encoders~\cite{meng2025vlm2vecv2}, motivated by the robust associations these models can establish between spatial pixel regions and semantic tokens~\cite{bordes2024intro}.

\begin{figure}[t]
  \centering
   \setlength{\abovecaptionskip}{6pt}
   \includegraphics[width=0.99\linewidth]{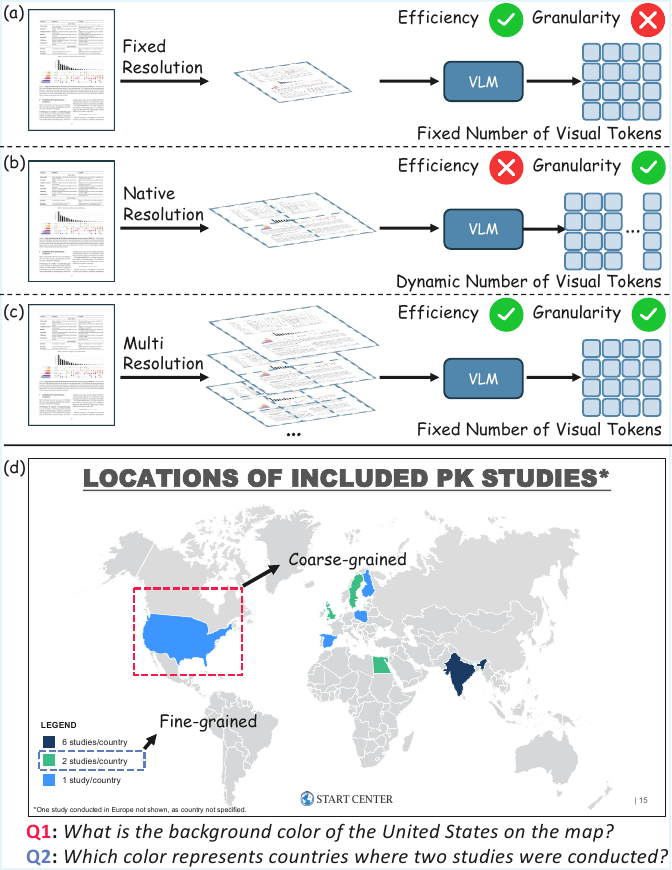} 
  \caption{Comparison between our approach (c) and two existing methods (a) and (b). (d) is a sample that demonstrates the necessity of both coarse- and fine-grained features. }
  \vspace{-0.6cm}
  \label{fig:motivation_and_case}
\end{figure}

In pursuit of more powerful visual document representation, we rethink the spatial configuration of input images in existing VLM-based encoders for VDR, which handle the various image resolutions and aspect ratios in either a \emph{fixed-resolution} or a \emph{native-resolution} manner~\cite{wang2024qwen2,dehghani2023patch}. 
As shown in Figure \ref{fig:motivation_and_case} (a), \emph{fixed-resolution} methods~\cite{liu2024improved,fayssecolpali} resize an input image to a fixed resolution (\eg $336 \times 336$ pixels) before encoding.
Though computationally efficient, these models often fail to obtain the intricate, fine-grained nuances inherent in complex, information-dense document images~\cite{ye2023ureader}.
In contrast, \emph{native-resolution} methods~\cite{liu2024llavanext, wang2025internvl3} (see Figure \ref{fig:motivation_and_case} (b)) preserve the original resolution or aspect ratio of input images, gaining better perception of fine-grained visual features.
However, they suffer from substantial indexing storage and computational overhead, due to the dynamic and excessive generation of visual tokens, which is often exacerbated for high-resolution images~\cite{niu2025native}.
Moreover, both approaches are constrained to \textbf{a single, static perspective}, partitioning images into a fixed grid of patches and failing to capture high-level layout structure and low-level granular details simultaneously within document images.

Our objective is to develop a method that can effectively and efficiently integrate coarse- and fine-grained visual features into a unified document image representation.
Take Figure~\ref{fig:motivation_and_case} (d) as an example.
Given Q1 (\ie \textit{``What is the background color of the United States on the map?''}), a model has to adopt a coarse-grained, high-level perspective to locate the region of the United States and identify the background color.
Comparably, Q2 (\ie \textit{``Which color represents countries where two studies were conducted?''})  demands a shift to a fine-grained perspective to extract the correct color cue from the map's legend.
To capture multi-scale perspectives, as illustrated in Figure~\ref{fig:motivation_and_case} (c), we devise a \textbf{multi-resolution sampling strategy} that resizes document images into multiple dimensions and processes them via VLMs.
To ensure storage efficiency and low latency, we further explore distilling the generated representations into a fixed number of visual tokens.

\begin{figure}[t]
  \centering
   \setlength{\abovecaptionskip}{6pt}
  \includegraphics[width=0.99\linewidth]{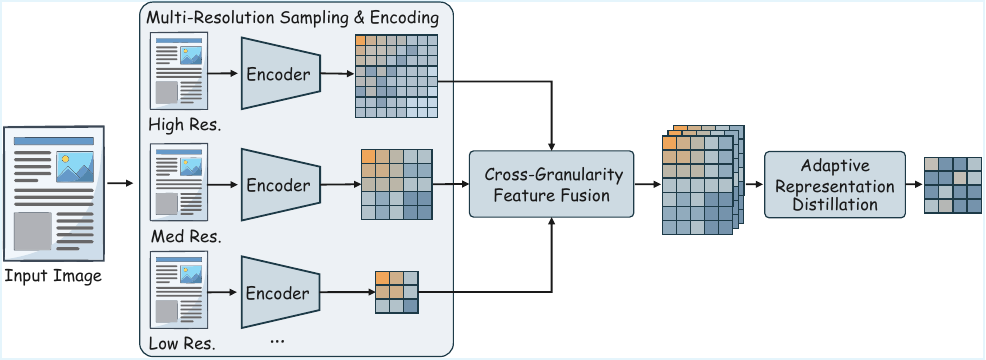} 
  \caption{An overview of our proposed \paradigm paradigm.}
  \vspace{-0.7cm}
  \label{fig:paradigm}
\end{figure}

\begin{figure}[t]
   \setlength{\abovecaptionskip}{6pt}
  \centering
  \begin{subfigure}[b]{0.43\linewidth} 
    \centering
    \includegraphics[width=\linewidth]{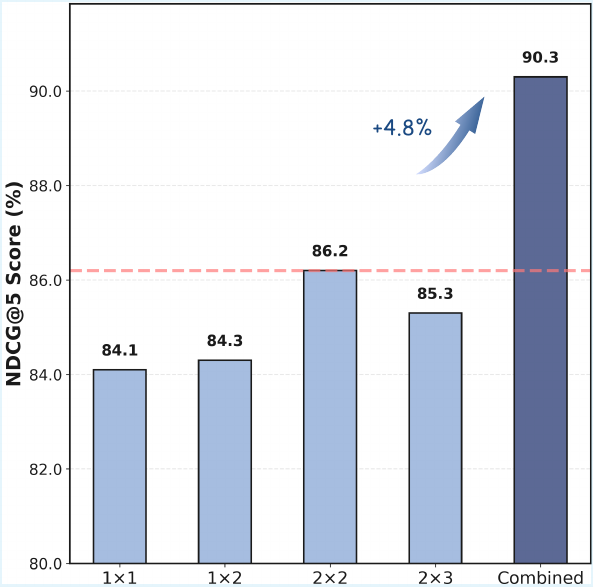} 
    \caption{\textbf{ViDoRe V1}}
    \label{fig:prelim_v1}
  \end{subfigure}
  \hfill 
  \begin{subfigure}[b]{0.43\linewidth}
    \centering
    \includegraphics[width=\linewidth]{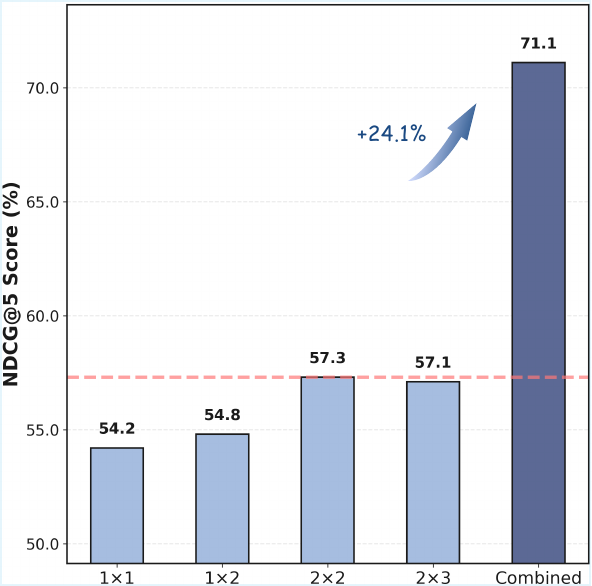} 
    \caption{\textbf{ViDoRe V2}}
    \label{fig:prelim_v2}
  \end{subfigure}
  \caption{\textbf{Preliminary experimental results.} 
    }
    \label{fig:preliminary}
    \vspace{-0.7cm}
\end{figure}

To verify its effectiveness, we conduct a preliminary study using an idealized combination. 
Specifically, we resize document images and partition them into $1\times1$, $1\times2$, $2\times2$, and $2\times3$ sampling grids to construct a pool of multi-granular representations. 
By dynamically routing each query to the granularity that yields the optimal result, we simulate the perceptual potential of an adaptive VDR model.
The preliminary studies on both ViDoRe V1~\cite{fayssecolpali} and ViDoRe V2~\cite{macé2025vidorebenchmarkv2raising} datasets with ColPali model~\cite{fayssecolpali} (see Figure~\ref{fig:preliminary}) reveal that:
1) On ViDoRe V1, the combined model outperforms the best single-resolution approach by +$\textbf{4.8}$\% relative gain in the NDCG@5 metric. This margin increases dramatically on ViDoRe V2, where the relative performance gap reaches +$\textbf{24.1}$\%.
2) The disparity suggests that integrating multi-scale perspectives is a promising approach to enhancing the performance and robustness of VDR models.

\begin{figure*}[t]
  \centering
   \setlength{\abovecaptionskip}{6pt}
   \setlength{\belowcaptionskip}{6pt}
  \includegraphics[width=\textwidth]{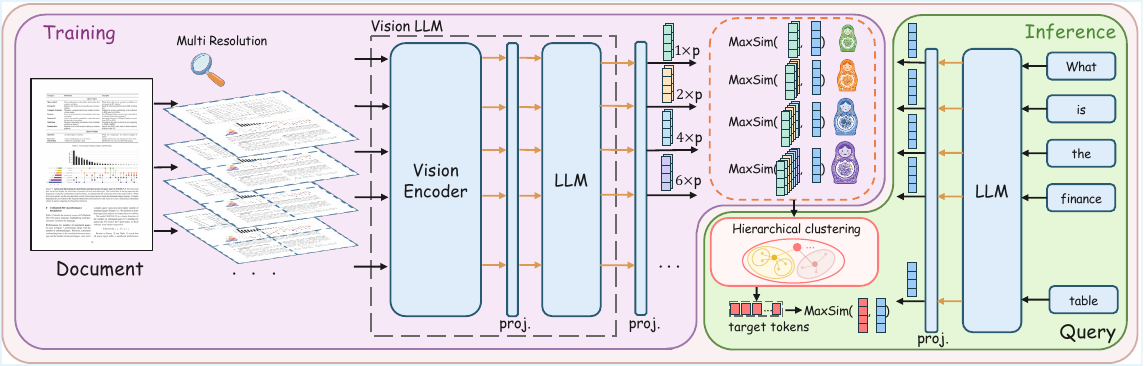}
  \caption{\textbf{The overall architecture of \method.} 
  The $S \times p$ represents the aggregate image patch token budget; $p$ denotes the \textbf{atomic token sequence} generated by the visual encoder from a single image pass; $S$ denotes the number of \textbf{spatial page divisions}.}
  \label{fig:main_model}
  \vspace{-0.5cm}
\end{figure*}

Based on these insights, we propose \textbf{\texttt{X-VisEmb}} (see Figure~\ref{fig:paradigm}), a new paradigm for effective and efficient document image encoding, which advances \textbf{from \emph{multi-resolution sampling and encoding}, through \emph{cross-granularity feature fusion}, to \emph{adaptive representation distillation}}. Accordingly, we develop \textbf{\textsc{MURE}} (see Figure~\ref{fig:main_model}), a novel framework designed to bridge the gap between coarse-to-fine visual perception and efficient representation.
\method first employs a multi-resolution sampling strategy, analogous to an "optical zoom", to observe the document image at hierarchical scales from coarse to fine. 
These multi-scale inputs are subsequently encoded using a VLM.
To facilitate the synergistic integration of features across different granularities, \method develops a Resolution-level Matryoshka representation learning (RMRL)~\cite{kusupati2022matryoshka} algorithm, which nests the embeddings at each resolution level to produce a unified, hybrid-grained document image representation.
During inference, it adopts a semantic-aware hierarchical clustering mechanism for elastic deployment. 
This enables users to dynamically select the optimal representation granularity from the unified embedding based on specific computational constraints, facilitating a flexible balance between retrieval accuracy and storage overhead.

Experiments on ViDoRe V1 and V2 demonstrate that \method beats strong baselines with lower storage overhead and latency, highlighting the superior performance-efficiency trade-off of our approach. Specifically, \method sets a new SOTA among PaliGemma-based models, outperforming ColMate by \textbf{+1.9\%} and \textbf{+2.3\%} in NDCG@5 on ViDoRe V1 and V2, respectively. Notably, the 512-token variant significantly surpasses ColPali on both datasets, achieving a \textbf{50\% storage reduction} while maintaining superior performance.

Our major contributions are summarized at below:
\begin{itemize}[leftmargin=*]
    \item We introduce a novel \textbf{multi-resolution sampling and encoding strategy} for VDR tasks to simultaneously capture visual features at varying granularities and verify its feasibility via preliminary studies.
    \item We propose a novel \textbf{\paradigm} paradigm for effective and efficient document image encoding, advancing from \emph{multi-resolution sampling and encoding}, through \emph{cross-granularity feature fusion}, to \emph{adaptive representation distillation}.
    \item Built upon this new paradigm, we introduce \textbf{\method}, a novel VDR framework that utilizes a VLM as a hierarchical multi-resolution encoder, incorporates resolution-level Matryoshka Representation Learning for synergistic feature fusion, and employs a semantic-aware hierarchical clustering mechanism to facilitate representation distillation.
    \item Extensive experiments on two widely used VDR datasets show that our proposed \method consistently beats strong baselines, achieving SOTA performance; notably, it surpasses ColPali while using only 50\% of its visual tokens (\ie 512 tokens).
    Our method attains superior model performance with low storage cost and computational overhead, achieving a favorable effectiveness-efficiency trade-off.
\end{itemize}

\section{Preliminary}

\subsection{Task Definition}
Let $\mathcal{C} = \{doc_1, doc_2, \dots, doc_{|\mathcal{C}|}\}$ denote a collection of documents, where each document $doc_i$ comprises $P_i$ page images. 
With all documents $\mathcal{C}$, we aggregate a set of page images $\mathcal{P} = \bigcup_{i=1}^{|\mathcal{C}|} P_i = \{p_1, p_2, \dots, p_N\}$, where $p_j$ denotes an individual page image and $N$ represents the total page images.
Given a textual query $q$, a scoring function $s(q, p)$ is employed to quantify the semantic relevance between query $q$ and each page image $p \in \mathcal{P}$. 
The objective is to obtain the top-$K$ page images from $\mathcal{P}$ that are most relevant to $q$.

\subsection{Preliminary Study}
\label{sec:preliminary}
To validate our hypothesis that multi-resolution sampling benefits document image representation and thereby improves model performance, we conduct a preliminary study using a combined selector.
Specifically, we define a set of granularity configurations $\mathcal{G} = \{1\times1, 1\times2, 2\times2, 2\times3\}$, representing diverse aspect-ratio preserving grids. 
We respectively train a VDR model for each granularity $g \in \mathcal{G}$ on the ViDoRe V1~\cite{fayssecolpali} training set. 
During evaluation, we test these models on both ViDoRe V1~\cite{fayssecolpali} and ViDoRe V2~\cite{macé2025vidorebenchmarkv2raising} test sets.
Let $\mathcal{Q} = \{q_1, \dots, q_{|\mathcal{Q}|}\}$ denote a set of evaluation queries, where $S$ is the total number of queries.
For each query $q_i \in \mathcal{Q}$, we compute the model performance (\ie NDCG@5) under each granularity, yielding a set of scores $\{r_{i}^{1\times1}, r_{i}^{1\times2}, r_{i}^{2\times2},r_{i}^{2\times3}\}$. 
To simulate the perceptual potential of an adaptive system, we identify the granularity index that  maximizes the performance for $q_i$ via:
\begin{equation}
    \vspace{-0.15cm}
    r^*_i = \operatorname*{max}_{g \in \mathcal{G}} (r_{i}^{g}).
\end{equation}
The combined performance is computed based on their optimal scores $r_{i}^{*}$ across all queries.
We present the preliminary experiment results in Figure~\ref{fig:preliminary} and observe that the combination method consistently outperforms the strongest single-resolution baseline, yielding +$\textbf{4.8}$\%  and +$\textbf{24.1}$\%  relative improvement on ViDoRe V1 \cite{fayssecolpali} and  ViDoRe V2 \cite{macé2025vidorebenchmarkv2raising}, respectively.
It reveals that capturing document features across a hierarchical scale facilitates a more comprehensive visual representation, thereby significantly enhancing the performance of VDR models.

\section{\method Framework}

\subsection{\paradigm Paradigm}
As shown in Figure \ref{fig:paradigm}, \paradigm paradigm includes three key stages: 
\begin{itemize}[leftmargin=*]
\item \textbf{Multi-Resolution Sampling and Encoding}: To capture both high-level layout structure and low-level granular details, a multi-resolution sampling strategy is applied to resize the input image into a hierarchy of scales.
Then, the image at each scale is encoded into a high-dimensional latent space by a shared vision encoder.
\item \textbf{Cross-Granularity Feature Fusion}:
Subsequently, features from different resolutions are fused to capitalize on their complementary strengths, reconciling high-resolution local details with low-resolution global semantics.
By aligning these different granularities, the model constructs a holistic, hierarchical image representation that preserves structural integrity.
\item \textbf{Adaptive Representation Distillation}:
Lastly, the fused representations are adaptively compressed into a compact form, optimizing the trade-off between model performance and computational efficiency.  
This allows the \paradigm to maintain high efficacy across diverse downstream tasks while significantly reducing computational overhead and memory footprint.
\end{itemize}

Below, we will present \textbf{\method}, a novel framework designed following the \paradigm paradigm.
As illustrated in Figure \ref{fig:main_model}, \method employs a multi-resolution sampling strategy with predefined granularity levels and leverages the LLM attention mechanism with a resolution-level Matryoshka representation learning algorithm for cross-granularity feature fusion. 
At inference, \method applies a semantic-aware hierarchical clustering mechanism to achieve adaptive token compression.

\subsection{Multi-Resolution Sampling and Encoding}
\label{sec:sampling}

The \method framework employs a hierarchical multi-resolution sampling strategy to capture document semantics across varying scales. 
Let $\mathbf{I} \in \mathbb{R}^{H \times W}$ denote the input document image. 
A set of granularity levels is defined as $\mathcal{G} = \{g_1, g_2, \dots, g_L\}$, where each level $g_k$ corresponds to a specific grid layout of $h_k \times w_k$ (\eg $1\times1, 1\times2, \dots$). For a specific level $k$, the image $\mathbf{I}$ is spatially partitioned into $h_k \times w_k$ local regions. 
These regions are resized to a fixed resolution $(H_v, W_v)$ to satisfy the input constraints of the vision encoder, forming a batch of sub-images denoted as $\mathbf{P}^{(k)}$. 
Subsequently, a shared visual encoder $\Phi_{\text{vis}}(\cdot)$ (initialized from a VLM, \eg SigLIP-So400m~\cite{zhai2023sigmoid}) processes $\mathbf{P}^{(k)}$, followed by a vision-language adapter $\Phi_{\text{proj}}(\cdot)$ that maps the extracted features into the latent space. 
The final feature representation for the $k$-th scale is formally derived as:
\begin{equation}
    \mathbf{V}^{(k)} = \Phi_{\text{proj}}(\Phi_{\text{vis}}(\mathbf{P}^{(k)})) \in \mathbb{R}^{M_k \times d}
\end{equation}
where $d$ represents the hidden dimension, and $M_k$ denotes the total number of visual tokens, which scales linearly with the grid layout $h_k \times w_k$. 
The resulting multi-scale feature set $\mathcal{V} = \{\mathbf{V}^{(1)}, \dots, \mathbf{V}^{(L)}\}$ constitutes a visual pyramid, preserving both global layout structures and fine-grained local details.

\subsection{Cross-Granularity Feature Fusion}
\label{sec:fusion}

\subsubsection{Hierarchical Document Representation}

To synthesize the hierarchical visual information, the multi-granular feature set $\mathcal{V}$ is first aggregated into a unified input sequence. 
Specifically, the token maps from all granularity levels are flattened and concatenated along the sequence dimension:
\begin{equation}
    \mathbf{X} = \operatorname{Concat}(\mathbf{V}^{(1)}, \mathbf{V}^{(2)}, \dots, \mathbf{V}^{(L)})
\end{equation}

Subsequently, the concatenated sequence $\mathbf{X}$ is fed into the LLM backbone of the VLM, denoted as $\Phi_{\text{llm}}(\cdot)$. 
This module utilizes its inherent self-attention mechanism to effectively fuse the multi-scale features, capturing dependencies between global structure and local details. 
Following this, a linear projection layer $\Psi_{\text{proj}}(\cdot)$ is employed to compress the high-dimensional outputs into a compact embedding space. 
This step ensures that the dense semantic information is preserved while significantly reducing storage overhead. 
Formally, the final hierarchical representation $\mathbf{H}$ is derived as:
\begin{equation}
    \mathbf{H} = \Psi_{\text{proj}}(\Phi_{\text{llm}}(\mathbf{X})).
\end{equation}

\subsubsection{Resolution-Level Matryoshka Representation Learning}
To facilitate the synergistic integration of multi-granularity features, a Resolution-Level Matryoshka Representation Learning (\textbf{RMRL}) algorithm is introduced.
Given the unified hierarchical representation $\mathbf{H} \in \mathbb{R}^{T_{seq} \times d}$, where $T_{seq}$ denotes the total sequence length, $\mathbf{H}$ is logically partitioned into segments corresponding to the input granularity levels $\mathcal{G} = \{g_1, \dots, g_L\}$. 
Let $\mathbf{H}^{(k)}$ represent the subset of projected tokens in $\mathbf{H}$ that corresponds strictly to the $k$-th grid scale. 
To construct the Matryoshka structure, a nested representation sequence $\mathbf{D}^{(1)} \subset \mathbf{D}^{(2)} \subset \dots \subset \mathbf{D}^{(L)}$ is generated. 
The representation at level $k$, denoted as $\mathbf{D}^{(k)}$, is formed by concatenating all token subsets from the coarsest level up to level $k$:
\begin{equation}
    \mathbf{D}^{(k)} = \operatorname{Concat}(\mathbf{H}^{(1)}, \mathbf{H}^{(2)}, \dots, \mathbf{H}^{(k)}).
\end{equation}
In this hierarchy, $\mathbf{D}^{(1)}$ serves as the coarsest global image representation (derived from the $1\times1$ view). Subsequent levels $\mathbf{D}^{(k)}$ (where $k>1$) progressively incorporate finer visual cues, resulting in holistic image representations with increasingly higher granularity.

\subsection{Model Training}

\subsubsection{Late Interaction}
The query $q$ is encoded by VLMs to yield an embedding $\mathbf{Q} \in \mathbb{R}^{N_q \times d}$, where $N_q$ denotes the number of query tokens and $d$ represents the hidden dimension.
To quantify the relevance between the query and the document representation at a specific level $k$, the MaxSim operation~\cite{khattab2020colbert} is employed. 
Formally, the score is computed by aggregating the maximum similarity of each query token against the document tokens in $\mathbf{D}^{(k)}$:
\begin{equation}
    s(q, \mathbf{D}^{(k)}) = \sum_{i=1}^{N_q} \max_{j=1}^{|\mathbf{D}^{(k)}|} (\mathbf{q}_i \cdot (\mathbf{d}_j^{(k)})^\top)
\end{equation}
where $\mathbf{q}_i \in \mathbf{Q}$ and $\mathbf{d}_j^{(k)} \in \mathbf{D}^{(k)}$ represent the $i$-th query token and the $j$-th document token at granularity level $k$, respectively.

\subsubsection{Granularity-Weighted Matryoshka Training Objective}
To enforce simultaneous optimization across all granularity levels, a joint loss function is formulated based on the InfoNCE objective. 
Consider a training batch $\mathcal{B} = \{(q_i, p_i)\}_{i=1}^{B}$ containing $B$ query-page pairs. 
For a specific granularity level $k$, let $\mathbf{D}_i^{(k)}$ denote the nested representation of page $p_i$ at this level. 
The contrastive loss for the $i$-th query is computed as the negative log-likelihood of the positive pair $(q_i, \mathbf{D}_i^{(k)})$ against all other documents in the batch:
\begin{equation}
    \mathcal{L}_{\text{NCE}}^{(k)}(q_i) = -\log \frac{\exp(s(q_i, \mathbf{D}_i^{(k)}))}{\sum_{j=1}^{B} \exp(s(q_i, \mathbf{D}_j^{(k)}))}.
\end{equation}
Here, the set $\{ \mathbf{D}_j^{(k)} \mid j \neq i \}$ serves as in-batch negative samples.

The final training objective aggregates these losses across all defined granularity levels $L$:
\begin{equation}
    \mathcal{L}_{\text{total}} = \sum_{k=1}^{L} w_k \cdot \frac{1}{B} \sum_{i=1}^{B} \mathcal{L}_{\text{NCE}}^{(k)}(q_i)
\end{equation}
where $w_k$ represents a scalar hyperparameter that balances the contribution of the $k$-th scale to the total loss. 
The optimization of this joint objective promotes semantic synergy, ensuring that the model comprehends image content via a coarse-to-fine, multi-granularity perceptual process.

\begin{table*}[t]
    \centering
    \caption{Performance on ViDoRe V1 (In-Domain). Best/second-best results are marked in bold/underlined in the category.}
    \label{tab:main_results_v1}
    
    \vspace{-0.3cm}
    
    \renewcommand{\arraystretch}{0.8} 
    \resizebox{\textwidth}{!}{ 
    \begin{tabular}{l c cccccccccccc}
        \toprule
        \textbf{Model} & \textbf{Size} & ArxivQ & DocQ & InfoQ & TabF & TATQ & Shift & AI & Energy & Gov. & Health & \textbf{Avg.} \\
        \midrule
        \rowcolor{aliceblue} \multicolumn{13}{c}{\textit{Other LLM as Backbone}} \\
        One-Peace & 4B & 43.9 & 23.4 & 59.9 & 57.0 & 13.4 & 17.0 & 45.4 & 53.2 & 55.9 & 59.5 & 42.9 \\
        E5-V & 8B & 41.1 & 24.3 & 49.5 & 58.2 & 9.0 & 13.2 & 46.1 & 57.7 & 53.0 & 59.6 & 41.2 \\
        DSE & 4B & 78.1 & 45.8 & 82.0 & 79.2 & 49.0 & 69.8 & 96.8 & 92.6 & 92.0 & \underline{96.3} & 78.2 \\
        Omni-Embed & 3B & \underline{85.3} & \underline{59.2} & \underline{89.2} & \textbf{91.0} & \underline{69.7} & \underline{78.6} & \underline{98.1} & \underline{93.5} & \underline{95.4} & 95.8 & \underline{85.7} \\
        ColQwen2 & 2B & \textbf{88.0} & \textbf{61.5} & \textbf{92.5} & \underline{89.0} & \textbf{82.2} & \textbf{89.9} & \textbf{99.0} & \textbf{95.9} & \textbf{95.5} & \textbf{98.8} & \textbf{89.2} \\
        \midrule
        \rowcolor{aliceblue} \multicolumn{13}{c}{\textit{PaliGemma-3B as Backbone}} \\
        ColPali & 3B & 83.0 & 58.4 & 85.7 & 87.4 & 70.3 & 77.3 & 97.4 & \textbf{95.4} & \textbf{96.2} & 96.9 & 84.9 \\
        ColPali(Reproduced) & 3B & 82.2 & 59.6 & 84.2 & 88.9 & 68.1 & 77.8 & 95.7 & 94.6 & 94.8 & 95.1 & 84.1 \\
        ColMate-Pali & 3B & 83.6 & 57.5 & 84.1 & 87.6 & 74.0 & 79.8 & \underline{98.3} & 94.1 & \underline{95.3} & 96.6 & 85.1 \\
        \rowcolor{lightcyan} \textbf{MURE}$^{\mathbf{512}}$ & 3B & 84.3 & \underline{61.7} & 87.9 & \textbf{90.6} & 73.2 & 78.9 & 97.2 & 93.2 & 93.1 & 96.4 & 85.7 \\ 
        \rowcolor{lightcyan} \textbf{MURE}$^{\mathbf{1024}}$ & 3B & 84.2 & \textbf{62.8} & 88.0 & \underline{90.0} & 74.9 & \underline{81.5} & 97.6 & 94.6 & 94.1 & 96.3 & \underline{86.4} \\ 
        \rowcolor{lightcyan} \bf MURE$^{\mathbf{1536}}$& 3B & \underline{84.4} & 61.5 & \underline{88.9} & 89.6 & \underline{75.3} & 81.1 & \underline{98.3} & 94.0 & 93.1 & \textbf{97.2} & \underline{86.4} \\ 
        \rowcolor{lightcyan} \textbf{MURE}$^{\mathbf{Full}}$ & 3B & \textbf{84.6} & \underline{61.7} & \textbf{89.0} & 89.3 & \textbf{76.8} & \textbf{83.2} & \textbf{98.7} & \underline{95.2} & 94.4 & \underline{97.1} & \textbf{87.0} \\ 
        \bottomrule
    \end{tabular}
    }
\end{table*}

\begin{table*}[t]
    \centering
    \caption{Performance on ViDoRe V2 (Out-of-Domain). Best/second-best results are marked in bold/underlined in the category.}
    \label{tab:main_results_v2}
    
    \vspace{-0.3cm}
    
    \renewcommand{\arraystretch}{0.8} 
    \small 
    \setlength{\tabcolsep}{6pt} 
    
    \resizebox{\textwidth}{!}{ 
    \begin{tabular}{ll cccccccc}
        \toprule
        \textbf{Model} & \textbf{Size} & ESG$_{\text{Human}}$ & Eco$_{\text{Mul}}$ & Bio$_{\text{Mul}}$ & ESG$_{\text{Syn\_Mul}}$ & Bio & ESG$_{\text{Syn}}$ & Eco & \textbf{Avg.} \\
        \midrule
        \rowcolor{aliceblue} \multicolumn{10}{c}{\textit{Other LLM as Backbone}} \\
        VisRAG-Ret & 3B & 53.7 & 48.7 & 47.7 & 46.4 & 54.8 & 45.9 & 59.6 & 51.0 \\
        VLM2Vec & 7B & 33.9 & 42.0 & 29.7 & 38.4 & 38.8 & 36.7 & 51.4 & 38.7 \\
        GME & 7B & \textbf{65.8} & \underline{56.2} & 55.1 & \textbf{56.7} & \textbf{64.0} & 54.3 & \textbf{62.9} & \underline{59.3} \\
        mmE5 & 11B & 52.8 & 44.3 & 46.8 & 54.7 & 51.3 & \underline{55.1} & 48.6 & 50.5 \\
        MoCa-3B & 3B & \underline{63.3} & \textbf{57.3} & \textbf{59.8} & \underline{54.8} & \underline{62.5} & \textbf{58.3} & \underline{62.8} & \textbf{59.8} \\
        ColQwen2 & 2B & 62.2 & 53.2 & \underline{56.5} & 54.2 & 61.8 & 53.4 & 61.5 & 57.5 \\
        \midrule
        \rowcolor{aliceblue} \multicolumn{10}{c}{\textit{PaliGemma-3B as Backbone}} \\
        ColPali & 3B & 51.1 & 49.9 & 56.5 & \underline{55.7} & 59.7 & 57.0 & 51.6 & 54.5 \\
        ColPali(Reproduced)  & 3B   & 55.5 & 50.8 & 53.6 & 52.4 & 59.3 & 55.9 & 52.1 & 54.2 \\
        ColMate-Pali & 3B & 62.8 & \underline{54.1} & \textbf{59.3} & 53.4 & \textbf{60.9} & 54.1 & 55.9 & 57.2 \\
        \rowcolor{lightcyan} \textbf{MURE}$^{\mathbf{512}}$ & 3B & 58.0 & 49.8 & 55.1 & 52.6 & 58.5 & 56.9 & 54.3 & 55.1 \\
        \rowcolor{lightcyan} \textbf{MURE}$^{\mathbf{1024}}$ & 3B & 62.7 & 50.7 & 55.8 & 55.0 & 59.1 & \underline{60.4} & 56.0 & 57.1 \\
        \rowcolor{lightcyan} \textbf{MURE}$^{\mathbf{1536}}$ & 3B & \underline{66.5} & 52.3 & 56.3 & 55.3 & 59.2 & 56.9 & \textbf{57.5} & \underline{57.7} \\
        \rowcolor{lightcyan} \textbf{MURE}$^{\mathbf{Full}}$ & 3B & \textbf{67.9} & \textbf{54.5} & \underline{56.6} & \textbf{57.4} & \underline{60.4} & \textbf{62.4} & \underline{57.3} & \textbf{59.5} \\
        \bottomrule
    \end{tabular}
    }
\end{table*}
\subsection{Semantic-aware Hierarchical Clustering}
\label{sec:inference}

While the multi-resolution sampling strategy captures comprehensive details, it inevitably increases the total volume of visual tokens, imposing constraints on storage and retrieval latency.
To mitigate this, a semantic-aware hierarchical token clustering mechanism is incorporated to adaptively distill the page representation. Specifically, during the offline indexing phase, Hierarchical Agglomerative Clustering  (HAC) \cite{clavie2024reducing} is applied to compress redundant visual information based on cosine similarity. This process iteratively merges semantically similar token pairs until the total token count is reduced to a predefined budget $N_{t}$. Consequently, the final distilled page representation is retained as a compact matrix $\tilde{\mathbf{H}} \in \mathbb{R}^{N_{t} \times d}$, where each row consists of a cluster centroid. This approach effectively preserves the semantic essence of the document while satisfying strict computational and storage constraints.
During the online retrieval stage, the relevance score is computed by applying the MaxSim operation between the query embedding $\mathbf{Q}$ and $\tilde{\mathbf{H}}$, thereby significantly reducing computational overhead.

\section{Experiments}
\subsection{Experimental Setup} 
\subsubsection{Datasets}
We conduct experiments on the ViDoRe benchmark~\cite{fayssecolpali} to assess visual document retrieval capabilities.

\noindent\textbf{Training Data.} For supervised fine-tuning, we utilize the official ViDoRe training split, comprising 118k query-page pairs aggregated from diverse synthetic and public sources.

\noindent\textbf{Evaluation Data.} To ensure comprehensive evaluation, we employ two benchmarks:
1) \textbf{ViDoRe V1 (In-domain)}~\cite{fayssecolpali} test set, which serves as the standard for in-domain performance, encompassing 10 datasets across academic and real-world domains; and 2) \textbf{ViDoRe V2 (Out-of-domain)}~\cite{macé2025vidorebenchmarkv2raising} test set, which consists of 7 datasets featuring more generalized settings and multilingual subsets for a robust evaluation.

\subsubsection{Compared Methods}
To evaluate the effectiveness of our proposed \method method, we compare it against following VDR models:
\noindent\textbf{1) VDR models with PaliGemma as backbone:} As our direct architectural counterparts, we compare against ColPali~\cite{fayssecolpali}, ColMate-Pali~\cite{masry2025colmate}.
\noindent\textbf{2) VDR models with other LLM as backbone:} We include a diverse set of MLLM-based VDR with various backbones, including  One-Peace~\cite{wang2023one}, E5-V~\cite{jiang2024e5}, DSE~\cite{ma2024unifying}, Omni-Embed-Nemotron~\cite{xu2025omni}, VisRAG-Ret~\cite{yuvisrag}, VLM2Vec~\cite{jiang2025vlm2vec}, GME~\cite{zhang2024gme}, mmE5~\cite{chen2025mme5}, MoCa~\cite{chen2025moca}, and ColQwen2~\cite{fayssecolpali}.
To ensure a fair comparison, all PaliGemma-based VDR models are trained exclusively on ViDoRe V1, eliminating performance gains from additional data.
Results of other VDR models are sourced from their original papers.

\subsubsection{Evaluation Metrics}
The metric of Normalized Discounted Cumulative Gain (NDCG@K) is adopted to evaluate the retrieval quality. 
Consistent with standard evaluation protocols, average NDCG@5 is reported across all tasks in this work.

\subsubsection{Implementation Details}
We employ PaliGemma-3B as the backbone architecture. For parameter-efficient fine-tuning, we apply Low-Rank Adaptation (LoRA)~\cite{hu2022lora} exclusively to the transformer layers of the language model component, configured with a rank $r=32$ and a scaling factor $\alpha=32$. Our multi-granularity model uses four hierarchical grid configurations: $\{g_1, g_2, g_3, g_4\}$ crops, corresponding to \{$1\times1$, $1\times2$, $2\times2$, and $2\times3$\} spatial divisions. The multi-level contrastive loss is weighted with coefficients $\{1.0, 1.5, 2.0, 2.5\}$ for each respective level. The model is trained on the ViDoRe V1 dataset for 2 epochs. We optimize the model using a learning rate of $5 \times 10^{-5}$, with the contrastive loss temperature set to $\tau = 0.02$. A total batch size of $b=64$ is utilized without gradient accumulation. All experiments are implemented based on the ColPali codebase and conducted on a server with 8 NVIDIA A100 GPUs.

\subsection{Main Results}

\begin{figure}[t]
 \setlength{\abovecaptionskip}{6pt}
   \setlength{\belowcaptionskip}{2.5pt}
  \centering
  \includegraphics[width=0.9\linewidth]{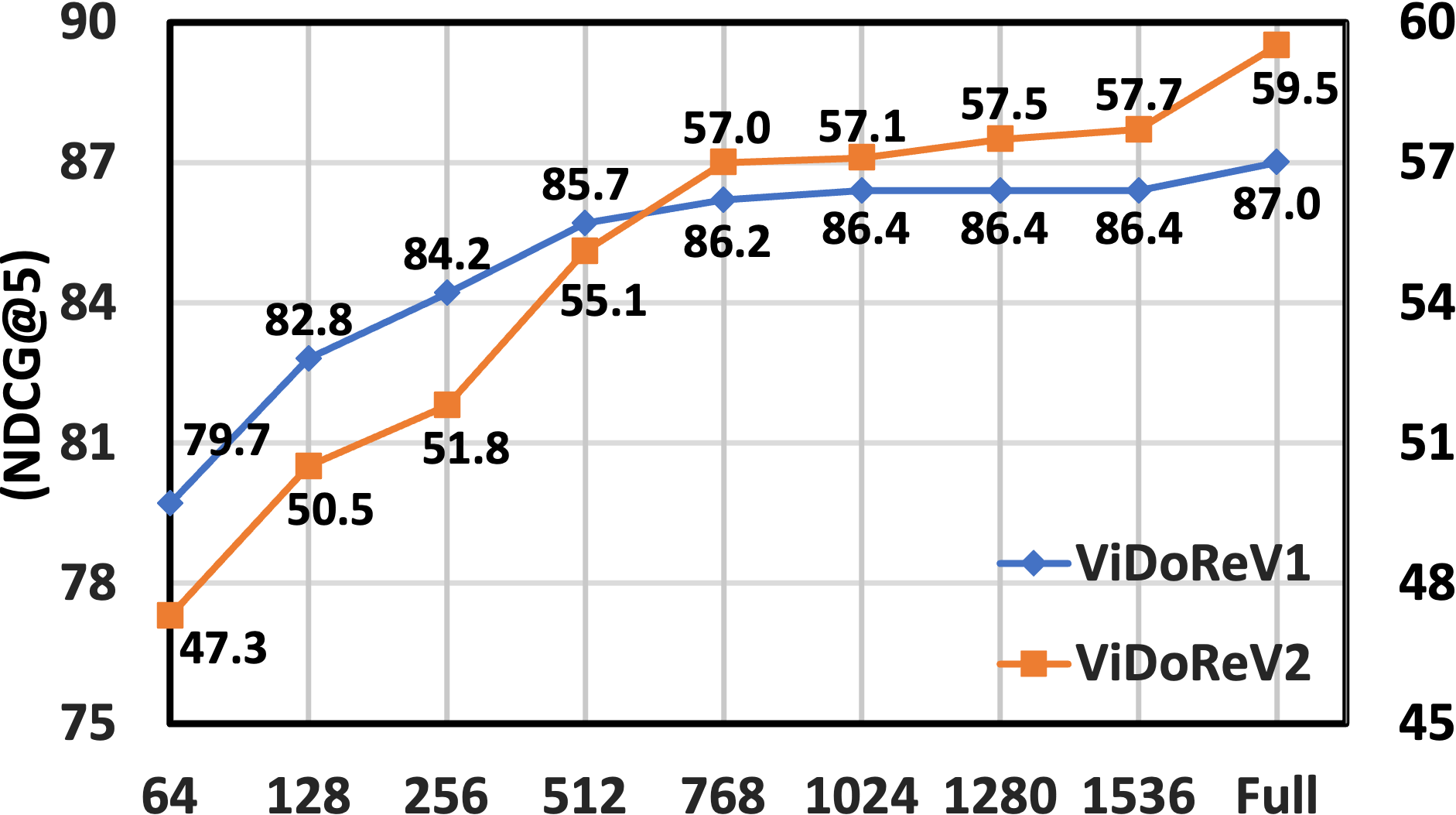} 
  \caption{\textbf{Performance analysis of \method with varying number of target token on ViDoRe V1 and ViDoRe V2.}}
  \vspace{-0.5cm}
  \label{fig:541}
\end{figure}

To evaluate the overall performance of our proposed \method framework against compared methods. 
We present the evaluation results on ViDoRe V1 and ViDoRe V2 benchmarks in Table~\ref{tab:main_results_v1} and Table~\ref{tab:main_results_v2}, respectively.
We can observe: 
1) \textbf{\method consistently outperforms all baselines within the PaliGemma-3B-based family on both datasets}.
For example, MURE$^{\textit{Full}}$ establishes a new SOTA among \textit{PaliGemma-3B} retrievers, surpassing ColMate by \textbf{+1.9\%} on ViDoRe V1 and \textbf{+2.3\%} on ViDoRe V2, respectively. Its superiority is particularly evident in information-dense scenarios, notably outperforming ColPali on \textit{TATQ} (\textbf{+6.5\%}) and \textit{Shift} (\textbf{+5.9\%}). These results confirm that our multi-resolution sampling strategy transcends the perceptual limits of fixed-resolution models, successfully capturing critical details. 
2) \textbf{\method demonstrates competitive performance compared to SOTA VDR models with alternative LLMs as backbones.}
Despite ColQwen2’s stronger backbone LLM and native-resolution support, MURE$^{\textit{Full}}$ still achieves a +2.0\% higher average score on ViDoRe V2 (out-of-domain), highlighting the effectiveness of the multi-resolution sampling strategy in handling diverse, unseen document structures.
3) \textbf{\method demonstrates superior robustness across various token budgets.} Specifically, at a 512-token budget, \method$^{512}$ already surpasses the full-resource ColPali by +0.8\% on ViDoRe V1 and +0.6\% on ViDoRe V2, effectively achieving a 50\% storage reduction while maintaining robust retrieval accuracy. Furthermore, under an identical 1,024-token constraint, \method$^{1024}$ significantly exceeds the top-tier ColMate by +1.3\% on ViDoRe V1 and remains competitive on ViDoRe V2. These results confirm that the multi-resolution sampling strategy and semantic-aware hierarchical clustering mechanism together ensure the robustness of \method with varying token budget. 
4) \textbf{Performance gap remains significant between \method and combination results reported in the preliminary study.} 
While \method achieves SOTA performance, there remains a clear margin of 3.3\% on ViDoRe V1 and 11.6\% on ViDoRe V2 compared to combined selector reported in the preliminary study \ref{sec:preliminary}. This gap suggests that further refining cross-granularity synergy is a promising research direction for unlocking the full potential of the \paradigm.

\subsection{In-Depth Analysis}

\subsubsection{Performance analysis with varying number of target tokens}

\begin{figure}[t]
 \setlength{\abovecaptionskip}{6pt}
   \setlength{\belowcaptionskip}{2.5pt}
  \centering
  \includegraphics[width=0.95\linewidth]{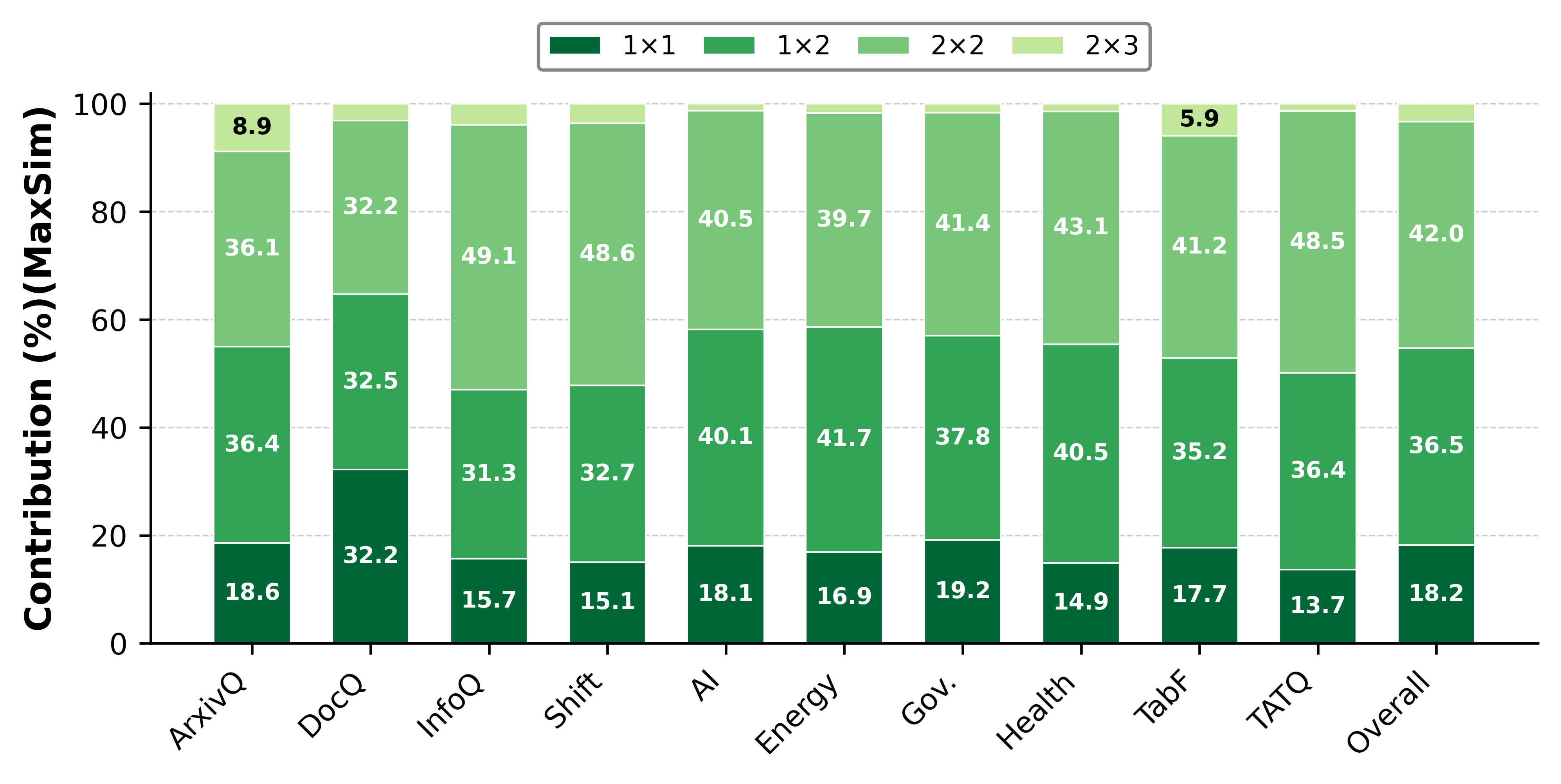} 
  \caption{\textbf{MaxSim score-weighted contribution analysis across all granularities.}}
  \vspace{-0.5cm}
  \label{fig:contribution}
\end{figure}

To investigate the trade-off between retrieval accuracy and storage efficiency, we evaluate the performance of \method across varying target token budgets $N_t \in \{64, 128, 256, 512, 768, 1024, 1280, 1536\}$. As shown in Figure~\ref{fig:541}, we make the following observations: 
1) \textbf{While higher token budgets generally correlate with better performance, we observe a clear trend of diminishing returns.} Specifically, increasing the budget from 64 to 512 yields a dramatic performance surge (+6.0\% on V1 and +7.8\% on V2), whereas further expanding the budget to 1536 results in only marginal gains (+0.7\% on V1 and +2.6\% on V2). This suggests that $512$ tokens serve as an efficiency ``sweet spot'' capturing the vast majority of discriminative semantics with significantly lower overhead. 
2) \textbf{\method demonstrates remarkable robustness under extreme compression}. With a tight budget of just 128 tokens, it preserves 95.2\% of the full-model performance on ViDoRe V1 (82.8\% vs. 87.0\%) and 84.9\% on ViDoRe V2 (50.5\% vs. 59.5\%). This indicates that our semantic-aware hierarchical clustering mechanism effectively prioritizes informative regions, allowing for orders-of-magnitude storage reduction with minimal degradation in retrieval accuracy.

\subsubsection{Contribution analysis of each granularity} 

To quantify the contribution of each image granularity to the final retrieval decision, we conduct a MaxSim score-weighted contribution analysis on ViDoRe V1. 
Specifically, we randomly select 100 queries for each dataset in ViDoRe V1 and perform the contribution analysis with the representation $\mathbf{D}^{(L)}$ obtained from Section ~\ref{sec:fusion}.
For each query token, we record its MaxSim score and the corresponding contributing document token.
To compute the contribution of each granularity to a query, we sum all scores within that granularity and normalize across granularities to obtain a contribution ratio.
We then average these ratios across all queries and present the results in Figure~\ref{fig:contribution}, from which we derive the following findings:
1) \textbf{Medium granularities dominate overall performance.}
The medium granularities (\ie $1\times2$ and $2\times2$) together account for \textbf{78.5\%} of the total contribution scores. This suggests that dividing a document image into 2 to 4 patches is likely to balance the capture of both local visual details and global semantic semantics.
\textbf{2) Granularity preference is highly task-dependent.}
The effectiveness of specific granularities varies according to the document characteristics. On visually complex datasets such as InfoQ and Shift, the model relies most heavily on the magnification capability of $g_3$ ($2\times2$), which contributes $49.1\%$ and $48.6\%$, respectively, to resolve intricate graphical elements like charts and slides. In contrast, the finer-grained $g_4$ ($2\times3$) contributes substantially less on both datasets.
Conversely, on the ArxivQA dataset, the contribution of $g_4$ increases to 8.87\%---nearly three times its average across datasets. This rise likely stems from the geometric alignment between the $2\times3$ grid and the landscape orientation of cropped academic paper figures, where finer partitioning becomes critical for recognizing small mathematical symbols.
\textbf{3) Synergy across all granularities is essential.}
Although medium granularities dominate the overall contribution, both the coarsest ($g_1$, $1\times1$) and finest ($g_4$, $2\times3$) granularities provide non-negligible and consistent contributions across all sub-datasets (e.g., $g1$ contributes $\sim$15.7\% on InfoQ; $g_4$ contributes 8.87\% on ArxivQA). This pattern confirms that the multi-granularity architecture functions as a visual zoom mechanism: $g_1$ preserves macroscopic context for maintaining global structure, while $g_4$ acts as a granular safeguard for capturing super-resolution details.

\subsection{Ablation Study}

\subsubsection{Effect analysis of each level in RMRL}
\begin{figure}[t]
 \setlength{\abovecaptionskip}{6pt}
   \setlength{\belowcaptionskip}{2.5pt}
  \centering
  \includegraphics[width=0.97\linewidth]{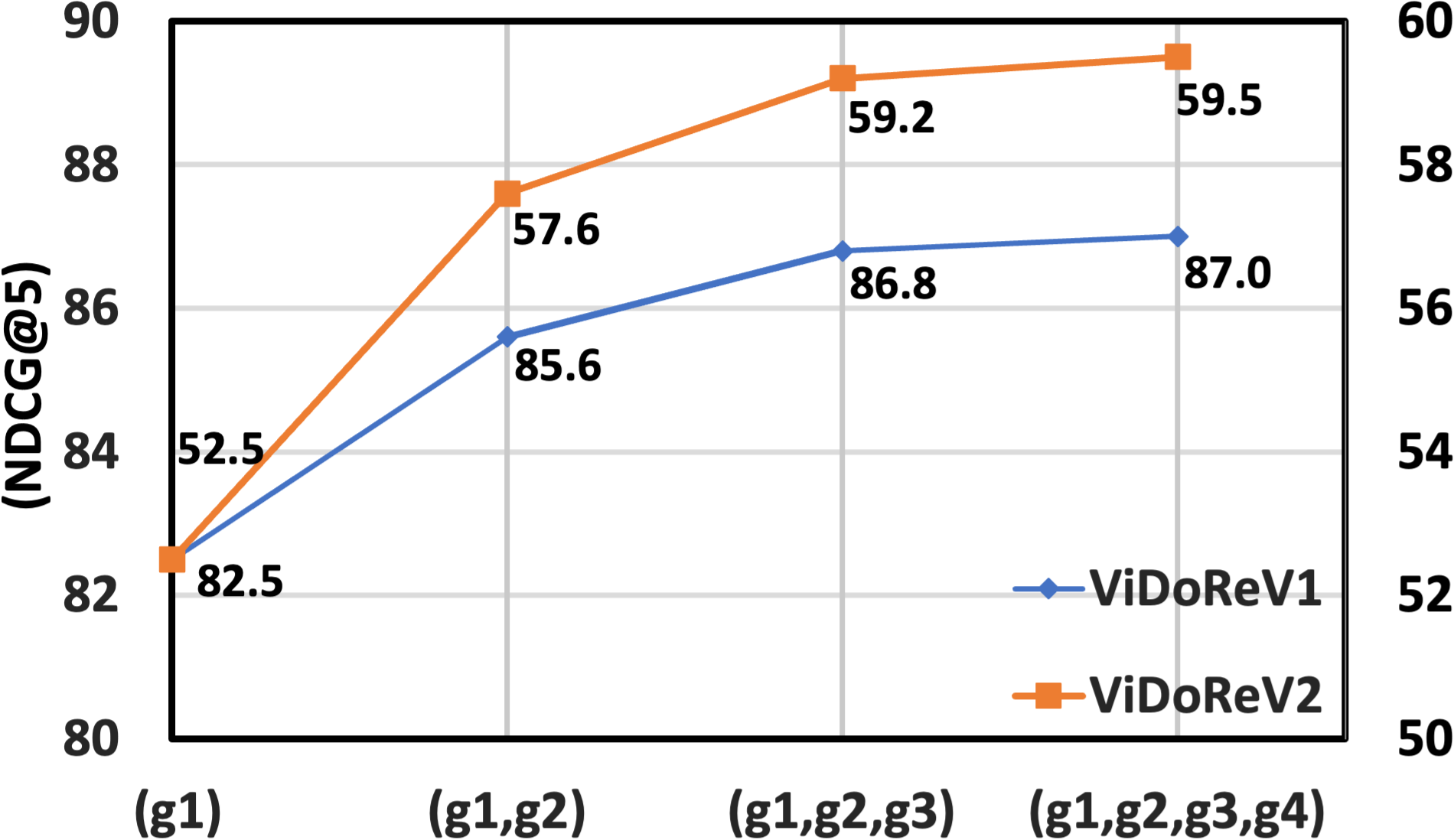} 
  \caption{Performance across nested granularity levels.}
  \vspace{-0.5cm}
  \label{fig:552}
\end{figure}

To investigate the effect of RMRL mechanism within the \method framework, we evaluate performance across all nested granularity levels.
Specifically, after training \method using the full grid configuration $\mathcal{G}=\{g_1, g_2, g_3, g_4\}$,  we progressively incorporate finer-grained scales during inference. The results, presented in Figure~\ref{fig:552}, reveal that expanding the nested levels yields consistent performance gains. Notably, extending the global view $(g1)$ to include ($g2$) yields a substantial boost (\ie +3.1\% on ViDoRe V1 and +5.1\% on ViDoRe V2). 
This upward trend continues as finer scales are added, culminating in peak performance with the full configuration. This confirms that our RMRL strategy effectively encodes cumulative visual details, allowing the model to leverage multi-granularity cues for precise retrieval.

\subsubsection{Performance comparison with different RMRL training strategies}
\begin{figure}[t]
 \setlength{\abovecaptionskip}{6pt}
  \centering
  \includegraphics[width=0.97\linewidth]{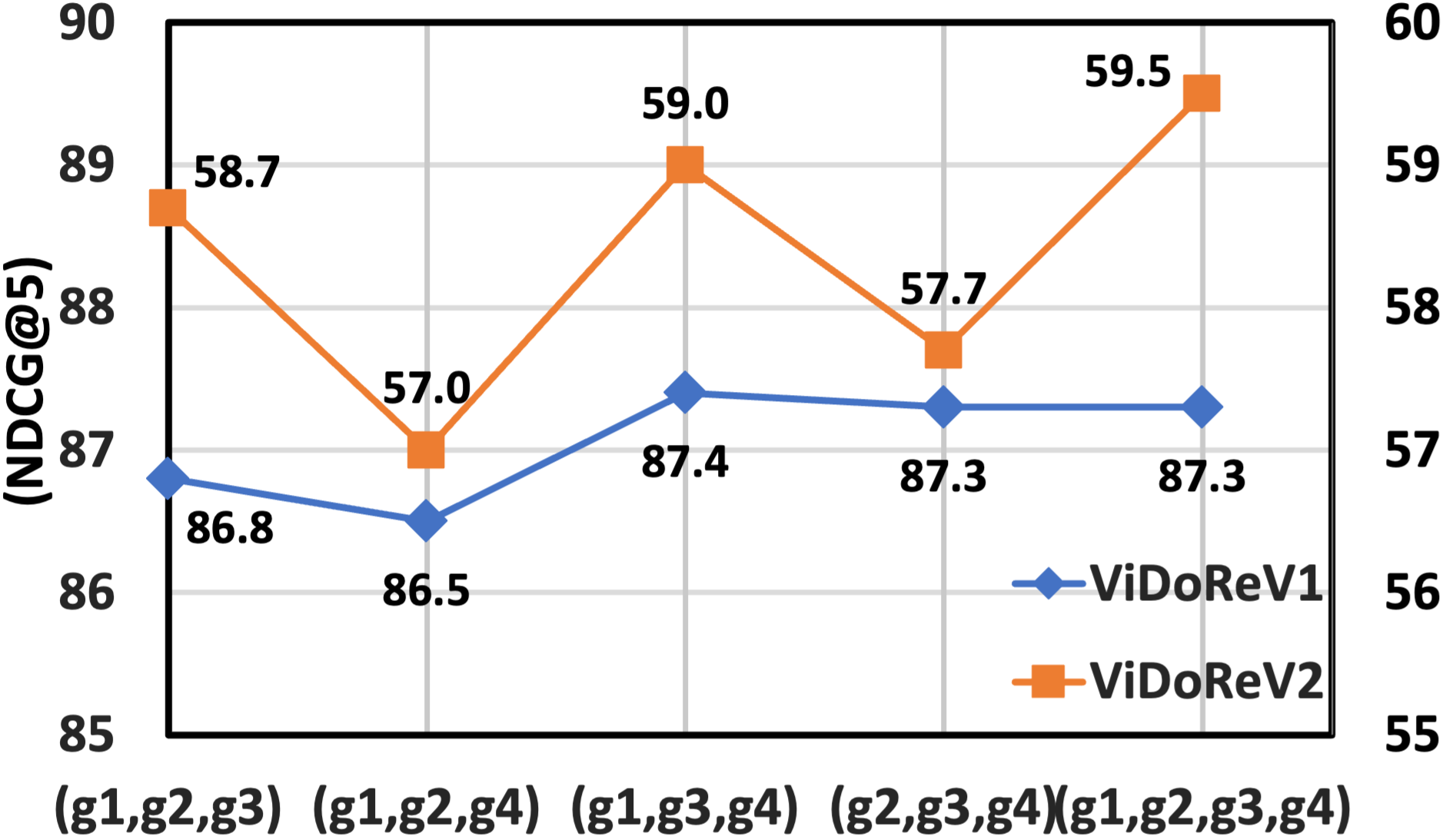} 
  \caption{Performance comparison using RMRL training with different granularity configurations.}
   \vspace{-0.6cm}
  \label{fig:553}
\end{figure}

To evaluate the contribution of each granularity within the RMRL in our proposed \method, we conduct a performance analysis by iteratively removing individual granularities from the full set. The experiments are performed across both the ViDoRe V1 and ViDoRe V2 benchmarks.
The results are summarized in Figure~\ref{fig:553}, from which we observe:
1) \textbf{The removal of any granularity results in a consistent degradation in performance on ViDoRe V2 (out-domain)}.
The full configuration establishes a clear lead over all other alternatives, indicating that each granularity in \method is crucial for enhancing model's generalization capabilities on complex, unseen documents.
2) \textbf{On the ViDoRe V1 (in-domain) benchmark, the configurations $(g_1,g_3,g_4)$ and $(g_2,g_3,g_4)$ achieve performance comparable to that of the full configuration.}
These result reveals that while adding granularities generally maintains performance, it does not always yield incremental gains, likely due to overfitting on in-domain datasets. Consequently, a promising research direction lies in identifying the optimal combination of granularities according to different document types.

\subsubsection{Performance analysis on different models with identical target tokens budget}

\begin{figure}[t]
 \setlength{\abovecaptionskip}{6pt}
   \setlength{\belowcaptionskip}{2.5pt}
  \centering
  \begin{subfigure}[b]{0.48\linewidth} 
    \centering
    \includegraphics[width=\linewidth]{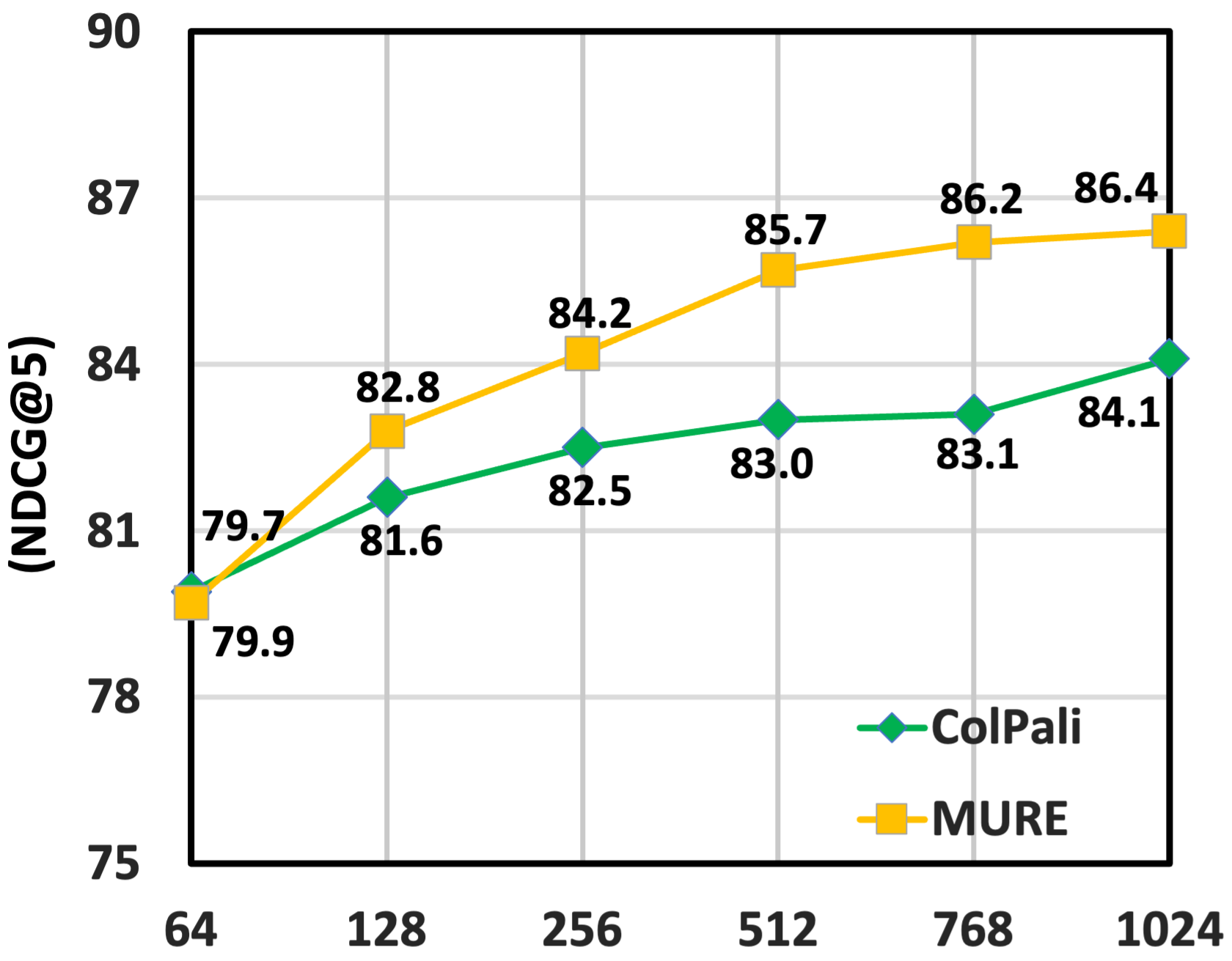} 
    \caption{\textbf{ViDoRe V1}}
    \label{fig:ablation3a}
  \end{subfigure}
  \hfill 
  \begin{subfigure}[b]{0.48\linewidth}
    \centering
    \includegraphics[width=\linewidth]{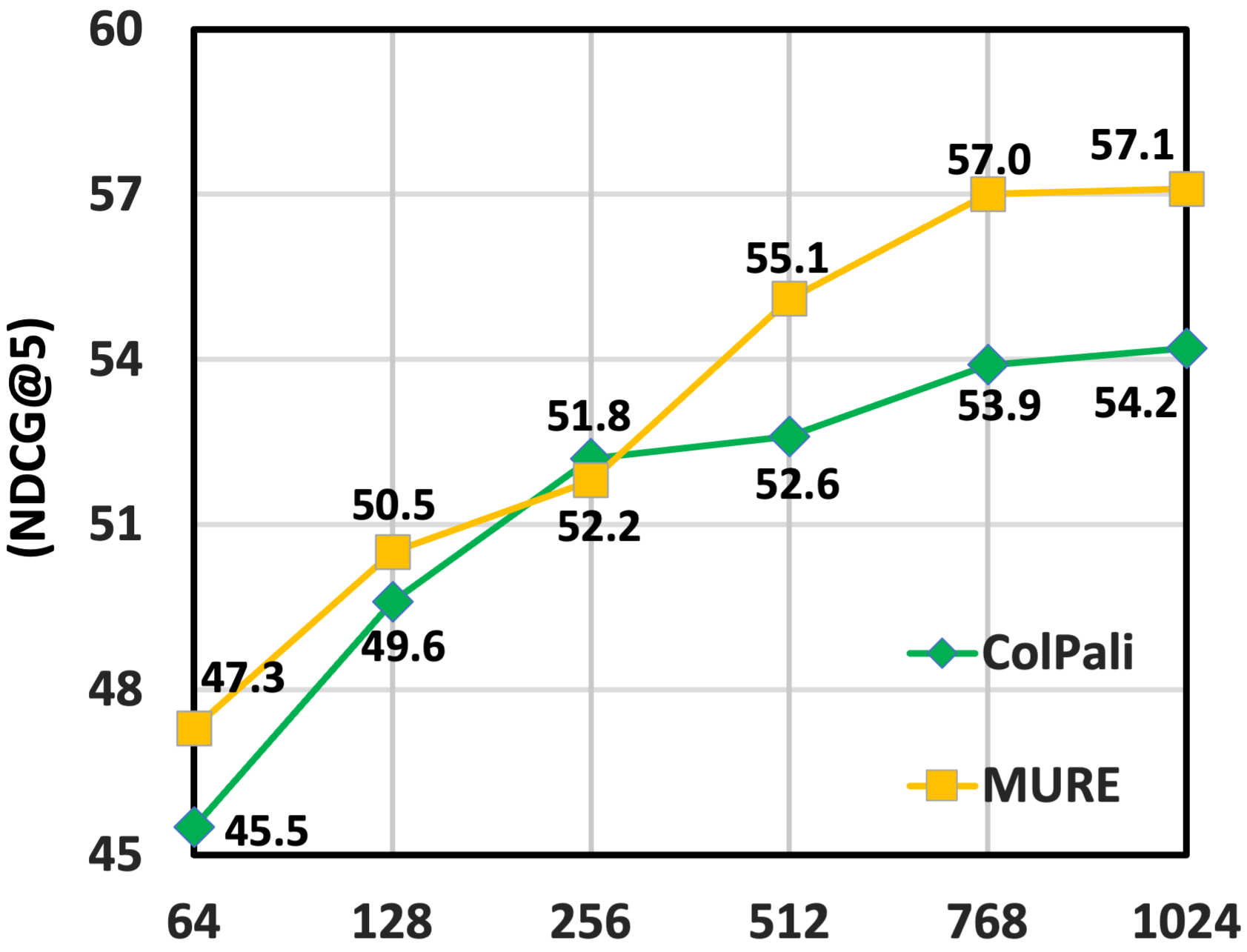} 
    \caption{\textbf{ViDoRe V2}}
   
    \label{fig:ablation3b}
  \end{subfigure}

  \caption{\textbf{Performance comparison between \method and ColPali across varying token budgets.} (a) Results on ViDoRe V1. (b) Results on ViDoRe V2.}
   \vspace{-0.7cm}
    \label{fig:551}
\end{figure}

To check the performance of different models under identical storage constraints, we compare our  \method with ColPali~\cite{fayssecolpali}.
Specifically, we conduct a comparative study by varying the target token budget $N_t \in \{64, 128, 256, 512, 768, 1024\}$. 
Both models are first trained on the ViDoRe V1 training set and then adopt the same semantic-aware hierarchical token clustering mechanism to obtain the same number of target tokens for evaluation.
The evaluation results on both ViDoRe V1 and ViDoRe V2 test sets, presented in Figure~\ref{fig:551}, reveal that: 1) \textbf{\method demonstrates consistent superiority over the ColPali across the entire spectrum of token budgets}. Specifically, at the standard budget of 1024 tokens, \method establishes a distinct advantage, outperforming the baseline by significant margins of \textbf{+1.5\%} on ViDoRe V1 and \textbf{+2.6\%} on ViDoRe V2. This confirms that our multi-granularity perception strategy captures significantly richer semantic information. 
2) \textbf{Performance gap remains significant under aggressive compression}. Even when restricted to a 512-token budget, \method consistently outperforms ColPali by 2.7\% on V1 and 2.5\% on V2. Furthermore, under extreme constraints (e.g., 128 tokens), \method maintains a clear margin of 1.2\% on V1 and 0.9\% on V2. These results validate the superiority and robustness of our \method, showcasing that it preserves critical information more effectively than baselines under significant token compression.

\section{Related Work}

\subsection{Visual Document Retrieval}
Visual Document Retrieval (VDR) models treat document pages as holistic visual representations, in contrast to traditional OCR-based pipelines that typically discard layout and visual cues~\cite{xu2020layoutlm,kim2022ocr}.
Current approaches fall into two main categories: \emph{Single-Vector VDR} and \emph{Multi-Vector VDR}. 
\emph{Single-Vector VDR} methods~\cite{ma2024unifying,yuvisrag,zhang2025qwen3} encode each document image into a fixed-size embedding through mean pooling or special tokens (\eg \texttt{[EOS]}), which often leads to the loss of fine-grained visual details demanded for precise matching ~\cite{fayssecolpali}.
To mitigate information compression, \emph{Multi-Vector VDR} methods\cite{fayssecolpali,masry2025colmate,cui2025attention,xiao2025metaembed} adopt the late-interaction mechanism and represent each document page as a sequence of patch embeddings to compute relevance via \texttt{MaxSim} operations~\cite{khattab2020colbert}.
They achieve superior accuracy, particularly on documents with high information density, but at the cost of substantially increased storage requirements and inference latency.
To alleviate these issues, visual token compression~\cite{ma2025towards,xiao2025metaembed,yao2025towards} has recently been explored to reduce index size by pruning or condensing redundant tokens. 
In this work, we propose a multi-vector VDR approach that leverages a multi-resolution VLM encoder and a hierarchical token compression strategy to achieve superior performance at substantially reduced storage costs compared to baseline approaches.

\subsection{Image Representation}
The encoding of images into a unified semantic space~\cite{tschannen2025siglip, li2023blip}, i.e. image representation, constitutes a cornerstone for a wide range of downstream tasks~\cite{mathew2021docvqa,zhu2022towards}.
With the rapid advancement of large vision–language models (VLMs), recent work has increasingly leveraged VLM-based architectures to learn such representations~\cite{wei2026deepseek,fayssecolpali,jian2025rzenembed}. 
Existing visual encoding strategies can be broadly classified into \emph{Fixed-Resolution}, \emph{Dynamic-Resolution} and \emph{Hybrid-Resolution} methods based on how they handle variations in image resolution and aspect ratio. 
\emph{Fixed-Resolution} methods, like LLaVA-1.5~\cite{liu2024improved} and PaliGemma~\cite{beyer2024paligemma}, resize the input image to a fixed dimension (\eg $336 \times 336$) prior to VLM encoding.
Despite their computational efficiency and structural simplicity, representations in this category often struggle to adequately capture the fine-grained nuances of complex, information-dense images~\cite{niu2025native, ye2023ureader}. 
\emph{Dynamic-Resolution} methods enhance the representation of fine-grained visual information by employing either adaptive image cropping~\cite{liu2024llavanext,wang2025internvl3,li2024monkey} or native resolution encoding~\cite{wang2024qwen2,dehghani2023patch}.
However, these approaches tend to suffer from a ``resolution dilemma'', as the highly variable and excessive number of visual tokens would lead to significant indexing storage and computational overhead~\cite{niu2025native, li2025tokenpacker}.
\emph{Hybrid-Resolution} methods\cite{Liu_2024_CVPR,tong2024cambrian} employ a dual-encoder strategy, combining a high-resolution encoder for local details with a low-resolution encoder for global context.
In this work, we propose a multi-resolution encoding paradigm, which produces a comprehensive coarse-to-fine representation through multiple sampling (\eg 4 different resolutions in this work) while maintaining a fixed number of visual tokens during inference.

\subsection{Matryoshka Representation Learning}
Matryoshka Representation Learning (MRL) \cite{kusupati2022matryoshka} applies a nested structure to high-dimensional vectors, explicitly imposing an order of importance across different representation granularities, facilitating elastic truncation to satisfy varying computational budgets.
It has been widely adopted in recent embedding models~\cite{zhang2024mgte, gunther2025jina, zhang2025qwen3}, and extended to the token sequence level.
For example, $M^3$~\cite{caimatryoshka}, MQT~\cite{hu2024matryoshka}, and MetaEmbed~\cite{xiao2025metaembed} leverage hierarchical spatial pooling or learnable queries to construct nested token sequences for flexible sequence management. 
In contrast to these token-level MRL, our work introduces what is, to the best of our knowledge, the first image-level MRL strategy. 
Our approach organizes visual information across multiple resolutions into a hierarchical structure, employing the MRL objective to facilitate nested learning across diverse granular layers.

\section{Conclusion and Future Work}

In this paper, we introduce a novel framework \method that captures both fine-grained visual details and global document structure for VDR while remaining computationally efficient. 
We address the core challenge in current VLM-based encoders by balancing perceptual depth with token efficiency with a new \paradigm paradigm. 
By integrating Resolution-level Matryoshka Representation Learning (RMRL) and semantic-aware hierarchical clustering, \method achieves a superior performance-efficiency trade-off compared to existing methods. 
Our experiments on two widely used VDR benchmarks demonstrate that the proposed multi-resolution sampling strategy effectively captures complementary visual cues across scales, allowing \method to consistently outperform strong baselines. 
Notably, our framework maintains competitive performance while utilizing only 50\% of the visual tokens, providing a highly scalable solution for large-scale document repositories.
 
Looking ahead, we aim to further advance the \paradigm paradigm in the following directions: 
1) \textbf{Generalization to broader modalities and tasks}, such as long-form video retrieval and end-to-end visual RAG systems, where capturing multi-scale context is essential for complex reasoning. 
2) \textbf{Refinement of feature fusion and compression}, exploring more sophisticated attention-based integration and learnable token condensation methods to further maximize information density. 
3) \textbf{Investigation of flexible multi-resolution sampling}, moving towards adaptive sampling strategies that dynamically optimize resolution and focus areas based on the document's layout complexity to achieve an even more refined balance between perceptual depth and computational cost.


\bibliographystyle{ACM-Reference-Format}
\bibliography{sample-base}

@article{zhang2025qwen3,
  title={Qwen3 Embedding: Advancing Text Embedding and Reranking Through Foundation Models},
  author={Zhang, Yanzhao and Li, Mingxin and Long, Dingkun and Zhang, Xin and Lin, Huan and Yang, Baosong and Xie, Pengjun and Yang, An and Liu, Dayiheng and Lin, Junyang and others},
  journal={arXiv preprint arXiv:2506.05176},
  year={2025}
}

@misc{bordes2024intro,
      title={An Introduction to Vision-Language Modeling}, 
      author={Florian Bordes and Richard Yuanzhe Pang and Anurag Ajay and Alexander C. Li and Adrien Bardes and Suzanne Petryk and Oscar Mañas and Zhiqiu Lin and Anas Mahmoud and Bargav Jayaraman and Mark Ibrahim and Melissa Hall and Yunyang Xiong and Jonathan Lebensold and Candace Ross and Srihari Jayakumar and Chuan Guo and Diane Bouchacourt and Haider Al-Tahan and Karthik Padthe and Vasu Sharma and Hu Xu and Xiaoqing Ellen Tan and Megan Richards and Samuel Lavoie and Pietro Astolfi and Reyhane Askari Hemmat and Jun Chen and Kushal Tirumala and Rim Assouel and Mazda Moayeri and Arjang Talattof and Kamalika Chaudhuri and Zechun Liu and Xilun Chen and Quentin Garrido and Karen Ullrich and Aishwarya Agrawal and Kate Saenko and Asli Celikyilmaz and Vikas Chandra},
      year={2024},
      eprint={2405.17247},
      archivePrefix={arXiv},
      primaryClass={cs.LG},
      url={https://arxiv.org/abs/2405.17247}, 
}

@InProceedings{Liu_2024_CVPR,
    author    = {Liu, Haotian and Li, Chunyuan and Li, Yuheng and Lee, Yong Jae},
    title     = {Improved Baselines with Visual Instruction Tuning},
    booktitle = {Proceedings of the IEEE/CVF Conference on Computer Vision and Pattern Recognition (CVPR)},
    month     = {June},
    year      = {2024},
    pages     = {26296-26306}
}

@inproceedings{
    tong2024cambrian,
    title={Cambrian-1: A Fully Open, Vision-Centric Exploration of Multimodal {LLM}s},
    author={Shengbang Tong and Ellis L Brown II and Penghao Wu and Sanghyun Woo and ADITHYA JAIRAM IYER and Sai Charitha Akula and Shusheng Yang and Jihan Yang and Manoj Middepogu and Ziteng Wang and Xichen Pan and Rob Fergus and Yann LeCun and Saining Xie},
    booktitle={The Thirty-eighth Annual Conference on Neural Information Processing Systems},
    year={2024},
    url={https://openreview.net/forum?id=Vi8AepAXGy}
}

@inproceedings{yuvisrag,
  title={VisRAG: Vision-based Retrieval-augmented Generation on Multi-modality Documents},
  author={Yu, Shi and Tang, Chaoyue and Xu, Bokai and Cui, Junbo and Ran, Junhao and Yan, Yukun and Liu, Zhenghao and Wang, Shuo and Han, Xu and Liu, Zhiyuan and others},
  booktitle={The Thirteenth International Conference on Learning Representations}
}

@misc{meng2025vlm2vecv2,
      title={VLM2Vec-V2: Advancing Multimodal Embedding for Videos, Images, and Visual Documents}, 
      author={Rui Meng and Ziyan Jiang and Ye Liu and Mingyi Su and Xinyi Yang and Yuepeng Fu and Can Qin and Zeyuan Chen and Ran Xu and Caiming Xiong and Yingbo Zhou and Wenhu Chen and Semih Yavuz},
      year={2025},
      eprint={2507.04590},
      archivePrefix={arXiv},
      primaryClass={cs.CV},
      url={https://arxiv.org/abs/2507.04590}, 
}

@inproceedings{zhu2022towards,
  title={Towards complex document understanding by discrete reasoning},
  author={Zhu, Fengbin and Lei, Wenqiang and Feng, Fuli and Wang, Chao and Zhang, Haozhou and Chua, Tat-Seng},
  booktitle={Proceedings of the 30th ACM International Conference on Multimedia},
  pages={4857--4866},
  year={2022}
}

@inproceedings{fayssecolpali,
  title={ColPali: Efficient Document Retrieval with Vision Language Models},
  author={Faysse, Manuel and Sibille, Hugues and Wu, Tony and Omrani, Bilel and Viaud, Gautier and HUDELOT, CELINE and Colombo, Pierre},
  booktitle={The Thirteenth International Conference on Learning Representations}
}

@article{xiao2025metaembed,
  title={Metaembed: Scaling multimodal retrieval at test-time with flexible late interaction},
  author={Xiao, Zilin and Ma, Qi and Gu, Mengting and Chen, Chun-cheng Jason and Chen, Xintao and Ordonez, Vicente and Mohan, Vijai},
  journal={arXiv preprint arXiv:2509.18095},
  year={2025}
}

@inproceedings{caimatryoshka,
  title={Matryoshka Multimodal Models},
  author={Cai, Mu and Yang, Jianwei and Gao, Jianfeng and Lee, Yong Jae},
  booktitle={The Thirteenth International Conference on Learning Representations}
}

@article{hu2024matryoshka,
  title={Matryoshka query transformer for large vision-language models},
  author={Hu, Wenbo and Dou, Zi-Yi and Li, Liunian and Kamath, Amita and Peng, Nanyun and Chang, Kai-Wei},
  journal={Advances in Neural Information Processing Systems},
  volume={37},
  pages={50168--50188},
  year={2024}
}

@article{kusupati2022matryoshka,
  title={Matryoshka representation learning},
  author={Kusupati, Aditya and Bhatt, Gantavya and Rege, Aniket and Wallingford, Matthew and Sinha, Aditya and Ramanujan, Vivek and Howard-Snyder, William and Chen, Kaifeng and Kakade, Sham and Jain, Prateek and others},
  journal={Advances in Neural Information Processing Systems},
  volume={35},
  pages={30233--30249},
  year={2022}
}

@misc{macé2025vidorebenchmarkv2raising,
      title={ViDoRe Benchmark V2: Raising the Bar for Visual Retrieval}, 
      author={Quentin Macé and António Loison and Manuel Faysse},
      year={2025},
      eprint={2505.17166},
      archivePrefix={arXiv},
      primaryClass={cs.IR},
      url={https://arxiv.org/abs/2505.17166}, 
}

@inproceedings{khattab2020colbert,
  title={Colbert: Efficient and effective passage search via contextualized late interaction over bert},
  author={Khattab, Omar and Zaharia, Matei},
  booktitle={Proceedings of the 43rd International ACM SIGIR conference on research and development in Information Retrieval},
  pages={39--48},
  year={2020}
}

@inproceedings{mathew2021docvqa,
  title={Docvqa: A dataset for vqa on document images},
  author={Mathew, Minesh and Karatzas, Dimosthenis and Jawahar, CV},
  booktitle={Proceedings of the IEEE/CVF winter conference on applications of computer vision},
  pages={2200--2209},
  year={2021}
}

@article{tschannen2025siglip,
  title={SigLIP 2: Multilingual Vision-Language Encoders with Improved Semantic Understanding},
  author={Tschannen, Michael and Gritsenko, Alexey and Wang, Xiao and Naeem, Muhammad Ferjad and Alabdulmohsin, Ibrahim and Parthasarathy, Nikhil and Evans, Talfan and Beyer, Lucas and Xia, Ye and Mustafa, Basil and others},
  journal={Localization, and Dense Features},
  volume={6},
  year={2025}
}

@inproceedings{zhai2023sigmoid,
  title={Sigmoid loss for language image pre-training},
  author={Zhai, Xiaohua and Mustafa, Basil and Kolesnikov, Alexander and Beyer, Lucas},
  booktitle={Proceedings of the IEEE/CVF international conference on computer vision},
  pages={11975--11986},
  year={2023}
}

@article{beyer2024paligemma,
  title={Paligemma: A versatile 3b vlm for transfer},
  author={Beyer, Lucas and Steiner, Andreas and Pinto, Andr{\'e} Susano and Kolesnikov, Alexander and Wang, Xiao and Salz, Daniel and Neumann, Maxim and Alabdulmohsin, Ibrahim and Tschannen, Michael and Bugliarello, Emanuele and others},
  journal={arXiv preprint arXiv:2407.07726},
  year={2024}
}

@article{wang2025internvl3,
  title={Internvl3. 5: Advancing open-source multimodal models in versatility, reasoning, and efficiency},
  author={Wang, Weiyun and Gao, Zhangwei and Gu, Lixin and Pu, Hengjun and Cui, Long and Wei, Xingguang and Liu, Zhaoyang and Jing, Linglin and Ye, Shenglong and Shao, Jie and others},
  journal={arXiv preprint arXiv:2508.18265},
  year={2025}
}

@inproceedings{kim2022ocr,
  title={Ocr-free document understanding transformer},
  author={Kim, Geewook and Hong, Teakgyu and Yim, Moonbin and Nam, JeongYeon and Park, Jinyoung and Yim, Jinyeong and Hwang, Wonseok and Yun, Sangdoo and Han, Dongyoon and Park, Seunghyun},
  booktitle={European Conference on Computer Vision},
  pages={498--517},
  year={2022},
  organization={Springer}
}

@inproceedings{xu2020layoutlm,
  title={Layoutlm: Pre-training of text and layout for document image understanding},
  author={Xu, Yiheng and Li, Minghao and Cui, Lei and Huang, Shaohan and Wei, Furu and Zhou, Ming},
  booktitle={Proceedings of the 26th ACM SIGKDD international conference on knowledge discovery \& data mining},
  pages={1192--1200},
  year={2020}
}

@article{ma2025towards,
  title={Towards Storage-Efficient Visual Document Retrieval: An Empirical Study on Reducing Patch-Level Embeddings},
  author={Ma, Yubo and Li, Jinsong and Zang, Yuhang and Wu, Xiaobao and Dong, Xiaoyi and Zhang, Pan and Cao, Yuhang and Duan, Haodong and Wang, Jiaqi and Cao, Yixin and others},
  journal={arXiv preprint arXiv:2506.04997},
  year={2025}
}

@inproceedings{ma2024unifying,
  title={Unifying Multimodal Retrieval via Document Screenshot Embedding},
  author={Ma, Xueguang and Lin, Sheng-Chieh and Li, Minghan and Chen, Wenhu and Lin, Jimmy},
  booktitle={Proceedings of the 2024 Conference on Empirical Methods in Natural Language Processing},
  pages={6492--6505},
  year={2024}
}

@inproceedings{lee2023pix2struct,
  title={Pix2struct: Screenshot parsing as pretraining for visual language understanding},
  author={Lee, Kenton and Joshi, Mandar and Turc, Iulia Raluca and Hu, Hexiang and Liu, Fangyu and Eisenschlos, Julian Martin and Khandelwal, Urvashi and Shaw, Peter and Chang, Ming-Wei and Toutanova, Kristina},
  booktitle={International Conference on Machine Learning},
  pages={18893--18912},
  year={2023},
  organization={PMLR}
}

@article{yao2025towards,
  title={Towards Efficient Multimodal Large Language Models: A Survey on Token Compression},
  author={Yao, Linli and Xing, Long and Shi, Yang and Li, Sida and Liu, Yuanxin and Dong, Yuhao and Zhang, Yi-Fan and Li, Lei and Dong, Qingxiu and Dong, Xiaoyi and others},
  journal={Authorea Preprints},
  year={2025},
  publisher={Authorea}
}

@inproceedings{li2023blip,
  title={Blip-2: Bootstrapping language-image pre-training with frozen image encoders and large language models},
  author={Li, Junnan and Li, Dongxu and Savarese, Silvio and Hoi, Steven},
  booktitle={International conference on machine learning},
  pages={19730--19742},
  year={2023},
  organization={PMLR}
}

@inproceedings{liu2024improved,
  title={Improved baselines with visual instruction tuning},
  author={Liu, Haotian and Li, Chunyuan and Li, Yuheng and Lee, Yong Jae},
  booktitle={Proceedings of the IEEE/CVF conference on computer vision and pattern recognition},
  pages={26296--26306},
  year={2024}
}

@inproceedings{ye2023ureader,
  title={Ureader: Universal ocr-free visually-situated language understanding with multimodal large language model},
  author={Ye, Jiabo and Hu, Anwen and Xu, Haiyang and Ye, Qinghao and Yan, Ming and Xu, Guohai and Li, Chenliang and Tian, Junfeng and Qian, Qi and Zhang, Ji and others},
  booktitle={Findings of the Association for Computational Linguistics: EMNLP 2023},
  pages={2841--2858},
  year={2023}
}

@article{niu2025native,
  title={Native Visual Understanding: Resolving Resolution Dilemmas in Vision-Language Models},
  author={Niu, Junbo and Zheng, Yuanhong and Miao, Ziyang and Dong, Hejun and Ge, Chunjiang and Liang, Hao and Lu, Ma and Zeng, Bohan and Zheng, Qiahao and He, Conghui and others},
  journal={arXiv preprint arXiv:2506.12776},
  year={2025}
}

@misc{liu2024llavanext,
  title={Llavanext: Improved reasoning, ocr, and world knowledge},
  author={Liu, Haotian and Li, Chunyuan and Li, Yuheng and Li, Bo and Zhang, Yuanhan and Shen, Sheng and Lee, Yong Jae},
  year={2024}
}

@inproceedings{li2024monkey,
  title={Monkey: Image resolution and text label are important things for large multi-modal models},
  author={Li, Zhang and Yang, Biao and Liu, Qiang and Ma, Zhiyin and Zhang, Shuo and Yang, Jingxu and Sun, Yabo and Liu, Yuliang and Bai, Xiang},
  booktitle={proceedings of the IEEE/CVF conference on computer vision and pattern recognition},
  pages={26763--26773},
  year={2024}
}

@article{li2025tokenpacker,
  title={Tokenpacker: Efficient visual projector for multimodal llm},
  author={Li, Wentong and Yuan, Yuqian and Liu, Jian and Tang, Dongqi and Wang, Song and Qin, Jie and Zhu, Jianke and Zhang, Lei},
  journal={International Journal of Computer Vision},
  pages={1--19},
  year={2025},
  publisher={Springer}
}

@article{wang2024qwen2,
  title={Qwen2-vl: Enhancing vision-language model's perception of the world at any resolution},
  author={Wang, Peng and Bai, Shuai and Tan, Sinan and Wang, Shijie and Fan, Zhihao and Bai, Jinze and Chen, Keqin and Liu, Xuejing and Wang, Jialin and Ge, Wenbin and others},
  journal={arXiv preprint arXiv:2409.12191},
  year={2024}
}

@article{dehghani2023patch,
  title={Patch n’pack: Navit, a vision transformer for any aspect ratio and resolution},
  author={Dehghani, Mostafa and Mustafa, Basil and Djolonga, Josip and Heek, Jonathan and Minderer, Matthias and Caron, Mathilde and Steiner, Andreas and Puigcerver, Joan and Geirhos, Robert and Alabdulmohsin, Ibrahim M and others},
  journal={Advances in Neural Information Processing Systems},
  volume={36},
  pages={2252--2274},
  year={2023}
}

@inproceedings{gunther2025jina,
  title={jina-embeddings-v4: Universal embeddings for multimodal multilingual retrieval},
  author={G{\"u}nther, Michael and Sturua, Saba and Akram, Mohammad Kalim and Mohr, Isabelle and Ungureanu, Andrei and Wang, Bo and Eslami, Sedigheh and Martens, Scott and Werk, Maximilian and Wang, Nan and others},
  booktitle={Proceedings of the 5th Workshop on Multilingual Representation Learning (MRL 2025)},
  pages={531--550},
  year={2025}
}

@inproceedings{zhang2024mgte,
  title={mGTE: Generalized Long-Context Text Representation and Reranking Models for Multilingual Text Retrieval},
  author={Zhang, Xin and Zhang, Yanzhao and Long, Dingkun and Xie, Wen and Dai, Ziqi and Tang, Jialong and Lin, Huan and Yang, Baosong and Xie, Pengjun and Huang, Fei and others},
  booktitle={Proceedings of the 2024 Conference on Empirical Methods in Natural Language Processing: Industry Track},
  pages={1393--1412},
  year={2024}
}

@article{wei2026deepseek,
  title={DeepSeek-OCR 2: Visual Causal Flow},
  author={Wei, Haoran and Sun, Yaofeng and Li, Yukun},
  journal={arXiv preprint arXiv:2601.20552},
  year={2026}
}

@article{jian2025rzenembed,
  title={Rzenembed: Towards comprehensive multimodal retrieval},
  author={Jian, Weijian and Zhang, Yajun and Liang, Dawei and Xie, Chunyu and He, Yixiao and Leng, Dawei and Yin, Yuhui},
  journal={arXiv preprint arXiv:2510.27350},
  year={2025}
}

@inproceedings{masry2025colmate,
  title={ColMate: Contrastive late interaction and masked text for multimodal document retrieval},
  author={Masry, Ahmed and Thakkar, Megh and Bechard, Patrice and Madhusudhan, Sathwik Tejaswi and Awal, Rabiul and Mishra, Shambhavi and Suresh, Akshay Kalkunte and Daruru, Srivatsava and Hoque, Enamul and Gella, Spandana and others},
  booktitle={Proceedings of the 2025 Conference on Empirical Methods in Natural Language Processing: Industry Track},
  pages={2071--2080},
  year={2025}
}

@article{cui2025attention,
  title={Attention Grounded Enhancement for Visual Document Retrieval},
  author={Cui, Wanqing and Huang, Wei and Guo, Yazhi and Hu, Yibo and Jin, Meiguang and Ma, Junfeng and Bi, Keping},
  journal={arXiv preprint arXiv:2511.13415},
  year={2025}
}

@article{jiang2024e5,
  title={E5-v: Universal embeddings with multimodal large language models},
  author={Jiang, Ting and Song, Minghui and Zhang, Zihan and Huang, Haizhen and Deng, Weiwei and Sun, Feng and Zhang, Qi and Wang, Deqing and Zhuang, Fuzhen},
  journal={arXiv preprint arXiv:2407.12580},
  year={2024}
}

@article{wang2023one,
  title={One-peace: Exploring one general representation model toward unlimited modalities},
  author={Wang, Peng and Wang, Shijie and Lin, Junyang and Bai, Shuai and Zhou, Xiaohuan and Zhou, Jingren and Wang, Xinggang and Zhou, Chang},
  journal={arXiv preprint arXiv:2305.11172},
  year={2023}
}

@article{xu2025omni,
  title={Omni-Embed-Nemotron: A Unified Multimodal Retrieval Model for Text, Image, Audio, and Video},
  author={Xu, Mengyao and Zhou, Wenfei and Babakhin, Yauhen and Moreira, Gabriel and Ak, Ronay and Osmulski, Radek and Liu, Bo and Oldridge, Even and Schifferer, Benedikt},
  journal={arXiv preprint arXiv:2510.03458},
  year={2025}
}

@inproceedings{jiang2025vlm2vec,
  title={VLM2Vec: Training Vision-Language Models for Massive Multimodal Embedding Tasks},
  author={Jiang, Ziyan and Meng, Rui and Yang, Xinyi and Yavuz, Semih and Zhou, Yingbo and Chen, Wenhu},
  booktitle={ICLR},
  year={2025}
}

@article{zhang2024gme,
  title={GME: Improving Universal Multimodal Retrieval by Multimodal LLMs},
  author={Zhang, Xin and Zhang, Yanzhao and Xie, Wen and Li, Mingxin and Dai, Ziqi and Long, Dingkun and Xie, Pengjun and Zhang, Meishan and Li, Wenjie and Zhang, Min},
  journal={arXiv preprint arXiv:2412.16855},
  year={2024}
}

@article{chen2025mme5,
  title={mme5: Improving multimodal multilingual embeddings via high-quality synthetic data},
  author={Chen, Haonan and Wang, Liang and Yang, Nan and Zhu, Yutao and Zhao, Ziliang and Wei, Furu and Dou, Zhicheng},
  journal={arXiv preprint arXiv:2502.08468},
  year={2025}
}

@article{chen2025moca,
  title={MoCa: Modality-aware Continual Pre-training Makes Better Bidirectional Multimodal Embeddings},
  author={Chen, Haonan and Liu, Hong and Luo, Yuping and Wang, Liang and Yang, Nan and Wei, Furu and Dou, Zhicheng},
  journal={arXiv preprint arXiv:2506.23115},
  year={2025}
}

@article{hu2022lora,
  title={Lora: Low-rank adaptation of large language models.},
  author={Hu, Edward J and Shen, Yelong and Wallis, Phillip and Allen-Zhu, Zeyuan and Li, Yuanzhi and Wang, Shean and Wang, Lu and Chen, Weizhu and others},
  journal={ICLR},
  volume={1},
  number={2},
  pages={3},
  year={2022}
}

@article{clavie2024reducing,
  title={Reducing the footprint of multi-vector retrieval with minimal performance impact via token pooling},
  author={Clavi{\'e}, Benjamin and Chaffin, Antoine and Adams, Griffin},
  journal={arXiv preprint arXiv:2409.14683},
  year={2024}
}

\end{document}